%% file: Ramanujan-Network-Initialization.tex
\documentclass{article} 
\usepackage{iclr2024_conference,times}
\usepackage{tikz, comment}
\usetikzlibrary{positioning,chains,fit,shapes,calc}
\definecolor{myblue}{RGB}{80,80,160}
\definecolor{mygreen}{RGB}{80,160,80}
\input{math_commands.tex}

\usepackage{amsfonts,amsmath}
\usepackage{booktabs}
\usepackage{hyperref}
\usepackage{url}
\usepackage{graphicx}
\usepackage{multirow}
\usepackage{rotating}
\usepackage{tikz}
\usepackage{hyperref}
\usetikzlibrary{positioning}

\newcommand{\bigzero}{\mbox{\normalfont\Large\bfseries 0}}
\newcounter{theorem}
\newtheorem{definition}[theorem]{Definition}

\title{Graph Expansion in Pruned Recurrent Neural Network Layers Preserve Performance}

\author{Suryam Arnav Kalra
\thanks{Corresponding author} \\
Department of Computer Science and Engineering\\
Indian Institute of Technology Kharagpur\\
Kharagpur, India. \\
\texttt{\{suryamkalra35\}@gmail.com} \\
\And
Arindam Biswas\\
Research Scientist\\
Polynom \\
Paris, France.\\
\texttt{\{arin.math\}@gmail.com} \\
\And
Pabitra Mitra\\
Department of Computer Science and Engineering\\
Indian Institute of Technology Kharagpur\\
Kharagpur, India.\\
\texttt{\{pabitra\}@cse.iitkgp.ac.in} \\
\And
Biswajit Basu\\
School of Engineering,\\
Trinity College Dublin\\
Dublin, Ireland.\\
\texttt{\{basub\}@tcd.ie} \\
}

\iclrfinalcopy 
\begin{document}

\maketitle

\begin{abstract}
Expansion property of a graph refers to its strong connectivity as well as sparseness. It has been reported that deep neural networks can be pruned to a high degree of sparsity while maintaining their performance. Such pruning is essential for performing real time sequence learning tasks using recurrent neural networks in resource constrained platforms. We prune recurrent networks such as RNNs and LSTMs, maintaining a large spectral gap of the underlying graphs and ensuring their layerwise expansion properties. We also study the time unfolded recurrent network graphs in terms of the properties of their bipartite layers. Experimental results for the benchmark sequence MNIST, CIFAR-10, and Google speech command data show that expander graph properties are key to preserving classification accuracy of RNN and LSTM. 
\end{abstract}

\section{Introduction}
\label{sec:introduction}
Analysis of Artificial Neural Networks (ANNs) following a connection base approach is a topical research direction as this not only mimics brain networks in neuroscience, but also can provide specific graph measures which can be used for analysis of performance and robustness of the networks.

Researchers in the recent years have explored if there is a relation between the functional aspects of an ANN and its graph structure, and if such a relation does exist then are there any characterization that explains the relationship between the structure of the graph and is performance \cite{lecun1998gradient, sermanet2013overfeat, zeiler2014visualizing, krizhevsky2017imagenet, simonyan2014very, he2016deep, szegedy2015going}. However, most of these studies have been performed on custom design architectures for specific tasks. Some researchers have also proposed ANNs with a better predictive performance by varying the connections (i.e., edges) between nodes and the operations they perform \cite{zoph2016neural, zoph2018learning, ying2019bench}. However, identifying measures which can quantify the influence of the structure on the performance for generic graph models of ANN is a matter of open research interest.


In spite of effectiveness and robustness of Deep Learning Networks (DNNs), there are still shortcomings due to limitations on resources for PCs and embedded systems, for implementation of these networks. DNNs put significantly high demand on computational power and energy consumption. With an aim to reduce the memory demand and also to accelerate the interface speed of neural networks pruning approaches are widely employed. It has been observed that many of these dense networks contain a subnetwork, termed as a `winning lottery ticket' which is sparse yet performs as well as the dense network \cite{frankle2018lottery}.

Recurrent neural networks and Long Short Term Memory (LSTM) are network structures with feedback connection that are useful in learning sequences. Computational requirements of such networks are often enormous, thus highlighting the need for extensive pruning and sparsification. 

Despite, several attempts being made, the most effective method to sparsify and prune networks is still Iterative Magnitude-based Pruning (IMP) \cite{imp}. This method is computationally intensive to run and is required to be rigorously executed for every different network. Recent advances in the research on neural networks have made pruning large feed forward and convolutional networks possible even with a small quantity of data leading to sample efficiency. Yet, when attempting to recover sparse recurrent networks, these techniques may not work suitably. The failure to achieve sparse recurrent networks may be either quantitative or qualitative in nature. Sometimes when more recent sophisticated pruning techniques are applied, RNNs typically obtain worse performance quantitatively than they do under a simple random pruning scheme. But, if the distribution of weights in a pruned Long Short Term Memory (LSTM) or RNN network tend to be concentrated in specific neurons and gates and not well dispersed across the entire architecture \cite {zhang19}, then it may lead to qualitative failure in the process of achieving a sparse network. Also, if the pruning leads to irregular network structure, then there are possibilities of degradation of performance/energy efficiency \cite{zhe18}. Network pruning based on regular graph structure has been proposed in \cite{chen23}.

It has been observed across a broad range of different types of neural network architectures (such as deep CNNs, stacked LSTM, and seq2seq LSTM models), that large-sparse models tend to consistently outperform small-dense models by achieving up to ten times reduction in number of non-zero parameters while preserving the accuracy \cite{zhu2017prune}. This provides a strong motivation to explore the basis of such observations and support or explain this by possible graph measures, if any. There would be possible benefits brought about from such analysis as there is an increasing trend to deploy deep neural networks at the edge for high-accuracy, applications involved are real-time data mining and user interaction, speech recognition and language understanding to name a few. The latter two often apply a DNN to encode an input sequence and then use a decoder to generate the output sequence.

Fully connected structures of RNNs  \cite{grossberg2013recurrent} and LSTMs \cite{hochreiter1997long} leading to the requirement of large memory footprint make these difficult to be implemented on resource limited hardware, though these recurrent networks are known to be useful for several applications as previously mentioned. Previous structured pruning methods can effectively reduce RNN size; however, how to find a good balance between high sparsity and high task accuracy, remains an open question. Failing to answer this would give us pruned models only, which lead to moderate speedup on custom hardware accelerators. 



In general, RNNs have a significant large number of parameters and hence the overall process of training involves enormous computational costs, as it is required to repeat the process over the recurrent structures for multiple time steps. To overcome this difficulty, the approach of RNN pruning has attracted increasing attention in recent years \cite{rnnpruning22}. It is beneficial leading to reduction of computational cost as the time step progresses. Nevertheless, most existing methods of RNN pruning are heuristic \cite{furuya22} and hence any quantitative measure to improve the performance would be a development forward.

Search of measures which can unravel the relationship between the structure and performance of the neural network has also been motivated by an important aspect of implementation of deep learning technologies. Lightweight design of learning models has become an important research direction in the application of deep learning.  Pruning or sparsification is an effective means to achieve a large reduction in model parameters and Floating Point Operations. The relation between the lottery ticket hypothesis and expander graph properties for fully connected and convolutional neural networks has been explored in \cite{pal22}. Similar graph measures that are easily computable can significantly assist in guiding the sparsification approach. 

In this paper, we generalise the study on expander properties of recurrent architectures, both theoretically and with experimental implementations. We show that the `layer-wise expansion’ property of a graph is crucial to maintain connectivity and flow of information. The criteria (based on bounds on spectral properties) that preserve the expansion property can be used to decide the adequacy of a sparsified RNN for lightweight architecture design. In fact, this approach in combination with other attributes can form the basis of a new cost-effective sparsification and pruning algorithms to be used for real-time applications. The claims on the `expander’ property-based technique for the analysis and testing of the suitability of a lightweight architecture have been verified on RNN and LSTM architectures (including the effect of noise) highlighting the Importance of different types of connectivity in these two cases. Additionally, the theoretical analysis of a bi-partite structure of the rolled out RNN graph leading to the spectral analysis of a Toeplitz matrix structure has provided additional insights for computation of the properties.




{\bf Contributions:} The contributions of the paper are the following:
\begin{itemize}
\item We propose an approach for analyzing generic RNN/LSTM architectures with respect to their graph expansion properties.
\item We have identified qualification criteria based on expansion properties for identifying suitable lightweight RNN/LSTM architecture with a view to real-time hardware applications.
\item We have experimentally observed that the RNN/LSTM architectures, even with highly reduced density satisfying the layerwise bipartite Ramanujan graph property for the rolled out network, represent resiliently connected global networks achieving high accuracy as compared to the original high-density networks.
\item Our approach for analysis and the proposed qualification criteria based on expansion properties for identifying suitable lightweight RNN architecture is robust to noise. This has been verified experimentally.
\item We have identified the importance of different types of connections at different layers in an RNN architecture; and hence this guides the decision on how to prune a dense RNN, if necessary.
\item The expander graph property used in our study may be used as a stopping criteria for network pruning algorithms for RNNs and LSTMs.
\end{itemize}

\section{Related Work}\label{sec:related}
Graph structure of neural networks has been previously studied in literature \cite{you20}. Several sparse neural network topology have been explored in this context for deep neural networks \cite{snn17}. Statistical network properties like the power law are seen to hold for such sparse networks \cite{feng2022power}. Recently, block sparse recurrent neural networks \cite{narang17} and sparse LSTM \cite{liu2019intrinsically} topologies have been suggested.

Sparsity in deep neural networks is often considered as a desirable property in the regularization framework \cite{thom13}. Sparsity may map to functional specializations \cite{gabriel21}. It is definitely beneficial from the point of view of fast inference in resource poor platforms like edge devices \cite{han2017ese}. It has been widely reported in literature that such sparse networks often retain performance of the original network \cite{frankle2018lottery, khona22, ding2021audio}. 

Connectivity of sparse neural network are recently being characterized by their spectral graph properties \cite{you20, mcdonald19, hod2021detecting}. Expander graph properties of deep fully connected and convolutional networks (CNN0 has been first reported in \cite{prabhu18}. It was established in \cite{pal22} that Ramanujan graph properties of sparse feedforward and CNN is essential for maintaining their performance. Graph connectivity property of recurrent neural networks and LSTMs has not been studied in literature so far.

Pruning RNNs has been explored for a long time in literature \cite{giles1994pruning}. Large implicit depth in RNN necessaites the pruning. Similarly, pruning is also useful in reducing computational complexity of LSTMs \cite{zhu2017prune}. Among the pruning methods stagewise magnitude based prunign is among the most widely used technique \cite{guiying22}. Chatzikonstantinou et al. proposed a novel RNN pruning method that considers the RNN weight matrices as collections of time-evolving signals \cite{chatzikonstantinou2021recurrent}. Spectral pruning algorithms that consider graph properties of the RNN structure are proposed very recently in \cite{furuya22}. Regular graph structure of RNN has been explored in \cite{chen23}. A one-shot pruning approach based on spectral properties of a RNN and having reduced computational effort as compared to iterative pruning has been proposed in \cite{zhang19}. Differential pruning of various RNN layers has been proposed in \cite{rnnpruning22}. Pruning algorithms based on supermasks have also been studied for transformer networks \cite{choromanski22a}. 

RNN and LSTM are often considered to be hard to prune due to the recurrent structure of the underlying graph. No guideline exists in literature regarding the degree of pruning possible while ensuring prediction performance. We propose to use expander graph property as a criteria for this purpose. The recurrent networks are pruned to a point till the sparse connectivity properties are preserved.

\section{Expander Graphs and Ramanujan Graphs}\label{sec:network}
	In this section, we discuss various properties of expanders which will be pertinent for the rest of the article. 
	An expander graph is a sparse graph that has strong connectivity properties. The connectivity can be quantified in different ways giving rise to different notions of expanders such as vertex expanders, edge expanders and spectral expanders. These notions are interrelated. Recall that a graph $\Gamma = (V,E)$ is a tuple consisting of a vertex set $V$ and an edge set $E$ which is a subset of $V\times V$.
	\subsection{Combinatorial Expansion}
	\begin{definition}[Expander and Cheeger constant]
		A graph $\Gamma = (V,E)$ is said to be an $\epsilon$-vertex expander if for every non-empty subset $X \subset V$ with $|X|\leq \frac{|V|}{2}$, we have $\frac{|\delta(X)|}{|X|}\geq \epsilon$, where $\delta(X)$ denotes the outer vertex boundary of $X$ i.e., the set of vertices in $\Gamma$ which are connected to a vertex in $X$ but do not lie in $X$. The infimum as $X$ runs over all subsets of $V$ satisfying the conditions above is known as the vertex Cheeger constant and is denoted by $\mathfrak{h}(\Gamma)$.
	\end{definition}  
	
	Edge expanders and the edge Cheeger constant $\mathbf{h}(\Gamma)$ are defined similarly, where in place of the vertex boundary, we consider the edge boundary i.e., the set of edges which have one vertex in $X$ and the other outside of $X$. The vertex Cheeger constant $\mathfrak{h}(\Gamma)$ and the edge Cheeger constant $\mathbf{h}(\Gamma)$ are related by the following equivalence
	$$\frac{\mathfrak{h}(\Gamma)}{D} \leq \mathbf{h}(\Gamma) \leq \mathfrak{h}(\Gamma),$$
	where $D$ denotes the maximum degree of the graph (the degree of each vertex is the number of edges going out from the vertex).  The equivalence allows us to speak about vertex expansion and edge expansion interchangeably. Intuitively, given a graph with high vertex (or edge) Cheeger constant, it is more difficult to separate any subset of the vertices from the rest of the graph. This allows for free flow of information throughout the network which the graph modelises. In the literature, having a high Cheeger constant is also known as having high combinatorial expansion. 
 
 \subsection{Spectral Expansion}
	The notion of spectral expansion is a bit different from combinatorial expansion. Given a finite undirected graph $\Gamma$ the eigenvalues $\lambda_{n}\leq \cdots \leq\lambda_{1}$ of its adjacency matrix are all real and $\lambda_{1} \leq D$ with equality iff the graph is $D$-regular.
 A graph $\Gamma = (V,E)$ is said to be a spectral expander if the quantities $\lbrace |\lambda_{1}| - |\lambda_{2}|, |\lambda_{1}| - |\lambda_{k}| \rbrace $ are both bounded away from zero, where $ k = n-1$ if the graph is bipartite and $k=n$ otherwise.
 \subsection{Discrete Cheeger--Buser Inequality}
 Ideally, to ensure free flow information within the network, our goal is to ensure that the graphs which modelise the networks have high combinatorial expansion. This is achieved via the discrete Cheeger--Buser inequality discovered independently by \cite{dodziuk1984difference} and by \cite{alon1985lambda1}. The inequality states that 
 $$\frac{\mathbf{h}(\Gamma)^{2}}{2}\leq \alpha_{2}\leq 2\mathbf{h}(\Gamma),$$ where $\alpha_{2}$ denotes the second smallest eigenvalue of the normalised Laplacian matrix of $\Gamma$ and is related to the eigenvalues of the adjacency matrix via 
 $$\frac{\lambda_i}{D} \leq 1 - \alpha_i\leq \frac{\lambda_i}{d}\,\, \forall i = 1,2,\ldots, n. $$
See \cite{Chung2016} for details. From the above, it is easy to check that a high $|\lambda_{1}| - |\lambda_{2}|$ ensures a high $\mathbf{h}(\Gamma)$ and vice-versa.
 Thus, the two notions of expansion are inter-connected and every spectral expander remains a combinatorial expander. They are actually equivalent for some classes of graphs, for instance bipartite graphs (as the adjacency spectrum is symmetric about the origin), variants of algebraic graphs \cite{Breuillard2015, Biswas2019, BiswasSaha2021, BiswasSaha2022, BiswasSaha2023, BiswasSahaExpansion2021} etc.
 
	\subsection{Ramanujan graph bounds}
	A $d$-regular graph is said to be a Ramanujan graph if  $\max \lbrace |\lambda_{2}|, |\lambda_{k}| \rbrace \leq 2\sqrt{d-1}$. In the case of bipartite graphs, $\lambda_{k} = \lambda_{2}$, hence the previous expression reduces to $|\lambda_{2}| \leq 2\sqrt{d-1}$. For fixed degree, with the sizes of the graphs growing larger and larger, these are the best possible expanders, as given by the Alon-Bopanna bound.  We refer to Hoory--Linial--Wigderson \cite{hoory2006expander} for the details.
 
 When the graphs modelising the network are irregular (and possibly weighted), to guarantee large expansion, we use two closely related quantities for $d$. The combinatorial quantity $d_{avg}$ which is the average degree taking into account all vertices and the spectral quantity $\lambda_{1}$ which is the largest eigenvalue of the adjacency operator. The use of these quantities is justified by the work of Hoory \cite{Hoory2005} and result in extremal families. Further, they have the added advantage of being easy to compute. Using them, we consider the following expressions $\Delta_{R}, \Delta_{S}$ with           
\begin{align}\label{eq:pareto mle2}
 \Delta_{R} &= \frac{2\sqrt{d_{avg}-1} - \lambda_{2}}{\lambda_{2}} \\
 \Delta_{S} &= \frac{2\sqrt{\lambda_{1}-1} - \lambda_{2}}{\lambda_{2}} 
\end{align}
We recall that these bounds were also considered in \cite{pal22}.

\section{Network Structures of RNN and LSTM} Recurrent networks 
and LSTMs 
are cyclic structures where the output from previous time step state as well as the current time step input is used to compute the current output. The recurrent structure can be unfolded over time steps of the sequence to obtain a graph without cycles. In the unrolled the network the weights are copied over the time steps and only the state and the input values change. Thus the unfolded RNN/LSTM can be viewed as a very deep feedforward network. The network structures of unrolled RNNs and LSTMs can be modelised using bipartite graphs. Given a RNN or a LSTM, we consider the complete bipartite network through which the input data passes during inference. 



In these sort of architectures, the bipartite networks are of size $m\times m$ where $m$ denotes the number of vertices in the first layer. Consider the (weighted) adjacency matrix $B$ of this network (modelised by the graph $\Gamma$) i.e., if there is a weight $w$ from $i$ to $j$ in the graph $\Gamma$, we shall substitute $|w|$ for the corresponding entry in the matrix. We will use an iterated magnitude based pruning and check that at each step, the Ramanujan graph bound $\Delta_{S} > 0$ is maintained. This ensures that the network remains a strong combinatorial expander via the discrete Cheeger inequality. The consideration of layerwise expansion also has a global implication (for small unrolling).

\subsection{Representation of the time unfolded network}\label{sec;topelitz}
In this section we briefly discuss the eigenvalue bounds for a simplified version of the time unfolded network.
Let $\begin{pmatrix}
	\bigzero_{m,m} & B \\
	B^{T} & \bigzero_{m,m}\\
\end{pmatrix}$ be the adjacency matrix of the  bipartite graph of size $2m$ modeling the initial layer $W_{xh}$ of the neural network. Let $k$ denote the number of times the data loops through the unfolded neural network, i.e., the sequence length.
We consider a simplified version of the unrolled network consisting of the layers $W_{xh}$. This network is also bipartite and of size $(k+1)m$. Let us denote its weighted adjacency matrix by $A$. The eigenvalues of $A$ can be obtained using the eigenvalues of $B$ in certain cases. It is clear that $A$ is a $(k+1)m \times (k+1)m$ matrix and

$$A = \begin{pmatrix}
	\bigzero_{m,m} & B & \bigzero_{m,m} & \cdots & \bigzero_{m,m} \\
	B^{T} & \bigzero_{m,m} & B & \cdots & \bigzero_{m,m}\\
	\bigzero_{m,m} & B^{T} & \bigzero_{m,m} & \cdots & \bigzero_{m,m}\\
	\vdots & \ddots & \ddots &\ddots &\vdots\\
	\bigzero_{m,m} & \cdots & B^{T} &  \bigzero_{m,m} & B\\
	\bigzero_{m,m} & \cdots & \bigzero_{m,m} & B^{T} & \bigzero_{m,m}\\
\end{pmatrix}.$$
\newline
 $A$ is a block tridiagonal Toeplitz matrix. The eigenvalues of a tridiagonal Toeplitz matrix are well known. Closed form expressions for eigenvalues of general block tridiagonal Toeplitz matrices are unknown. However, eigenvalue approximations exist see \cite{Nakatsukasa_2012}. A particular case is when $B$ is symmetric i.e., $B =B^{T}$. For these type of network structures, the eigenvalues take the form, 

$$\left\lbrace2\lambda^{(B)}_{i}\cos \left(\frac{\pi j}{k+2}\right)\right\rbrace_{i=1 \ldots m, j = 1\ldots k+1}.$$
See \cite{Abd2020}. 

\subsection{Relation between spectral Gap and accuracy}
\label{sec:spectral_perf}
The hypothesis for our study is that the preservation of expander properties of the recurrent neural network and LSTM is related to performance of these networks. More precisely, we claim that the network classification accuracy is high for the pruned sparse networks when the values of $\Delta{R}$ and $\Delta{S}$ (Equation~\ref{eq:pareto mle2}) are positive. Further pruning leads to drastic drop in prediction accuracy.

\section{Experimental Results}\label{sec:result}

\subsection{Datasets}\label{sec:data}
We provide results for three benchmark sequential data, namely, the sequential MNIST, the sequential CIFAR10, and the Google speech command data. For the sequential MNIST data \cite{le2015simple}, each row of a $28\times 28$ MNIST digit grey-scale image is presented as input at a single time step, leading to a sequence of length $28$ for the entire image. The rows are presented in scanline order. The task is to classify the sequence into one of the ten classes. We have considered this data set as both long and short term dependencies characterize the classification task. We also present results for noisy MNIST sequence data. In the noisy data a Gaussian noise with zero mean and $\sigma$ variance is added to $p$ fraction of the sequence points (pixels). We considered $p=0.20$ and $\sigma = 0.15, 0.30, 0.45, 0.60$, respectively to study the effect of varying degree of noise.

The CIFAR-10 dataset consists of 60000 $32 \time 32$ colour images in 10 classes, with 6000 images per class. The sequential CIFAR-10 \cite{chang2017dilated} is constructed by a similar process in scanline order from the CIFAR-10 images. There are 50000 training images and 10000 test images. In a similar manner of the MNIST data, we study noisy CIFAR-10 images too.

The Google Speech Commands dataset \cite{warden2018speech} has 65000 one-second long utterances of 10 short words, by thousands of different people. It is an audio dataset, with a sampling rate of 16kHz, of 35 spoken words designed to help train and evaluate keyword spotting systems in the presence of background noise or unrelated speech. The keywords e.g., backward, bed, bird, cat, dog, down, eight, five, follow, forward, four, go, happy, house, learn, left, marvin, nine, no, off, on, one, right, seven, sheila, six, stop, three, tree, two, up, visual, wow, yes, zero, are commands used by a voice assistant. Mel-scale spectogram extracted from the raw signal are used as input features. The sequence length of the mel-scale input feature is $k=400$. If we set the sequence length to $k$, then the input is passed as a $16000/k$ size vector at each time step (and the number of time steps is $k$).  The dataset is used as a benchmark evaluate the performance of RNN/LSTM on long sequences.

\subsection{Setup}
Experiments are performed for both RNN and LSTM. The hyperparameters for the network are shown in Table~\ref{tab:hyperparam}. Dense output layers are considered for each network. Classification accuracy on a test data ($20$\%) is used as the performance measure of the network. While we perform experiments with both RNN and LSTM for the MNIST data, only the LSTM is studied for the CIFAR-10 and Google Speech Command datasets. This is because the accuracy of RNN is quite low for these datasets.

\begin{table}[ht]
\caption{Hyperparameters for the experiments.}
\label{tab:hyperparam}
\begin{center}
\begin{small}
\begin{tabular}{ll}
\toprule
Learning rate   & $0.001$\\
Training epochs   & $20$\\
Pruning epochs & $20$\\
Batch size   & $100$\\
Optimizer   & Adams\\
Initialization & Kaiming uniform\\
Activation & tanh\\
Loss & Cross entropy\\
Hidden nodes & 128\\

\bottomrule
\end{tabular}
\end{small}
\end{center}
\vskip -0.1in
\end{table}

\subsection{Results}

The goal of our experiments is to study the effect of preserving expander graph properties on the performance of sparse RNN and LSTM. Iterative magnitude pruning \cite{imp} over a number of epochs is performed to sparsify the networks. The weights between input to hidden layers, feedback layers, and hidden to output layers are represented as $W_{xh}$, $W_{hh}$, and $W_{hy}$ each representing a bipartite graph. We only prune the $W_{xh}$ and $W_{hh}$ layers, leaving out the dense $W_{hy}$ layer unchanged. The time unfolded networks effectively contains $k$ identical copies of the $W_{hh}$ bipartite graph, where $k$ is the input sequence length. 

Preservation of expander property is characterized by the spectral gap of the bipartite graphs of RNN layers. The spectral gap is obtained by Equation~\ref{eq:pareto mle2}. Spectral gaps for both unweighted ($\Delta_R$, $\Delta_S$ and weighted ($\Delta_S$) bipartite graph representation are reported.  We plot the variation in test set accuracy with the $q$\% remaining fraction of network weights for RNN and LSTM along with the spectral gaps $\Delta_S$ and $\Delta_R$ (Equation~\ref{eq:pareto mle2}). We present the following combination of results:
\begin{itemize}
\item Dataset: MNIST, CIFAR-10, Google speech command
\item Architecture: RNN (MNIST), LSTM (all)
\item Graph representation: Weighted, Unweighted
\item Layer: $W_{xh}$, $W_{hh}$
\item Spectral gap: $\Delta_S$ (weighted, unweighted), $\Delta_R$ (unweighted)
\end{itemize}

\begin{figure}[htbp]
\begin{center}
\centerline{\includegraphics[width=\columnwidth]{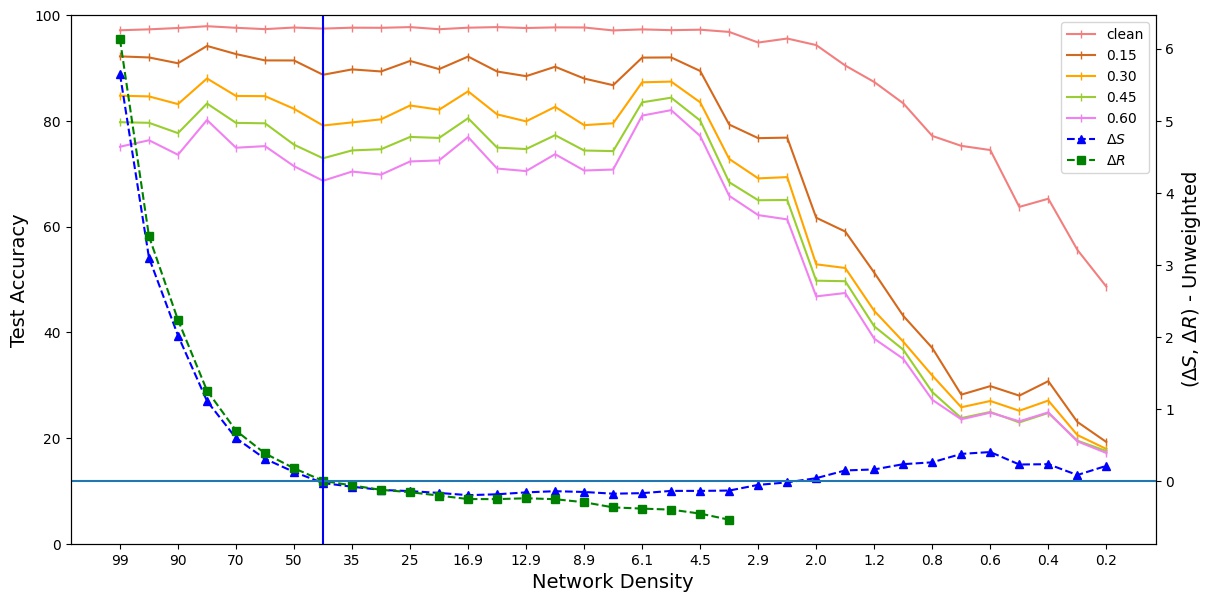}}
\centerline{(a) $W_{xh}$}
\centerline{\includegraphics[width=\columnwidth]{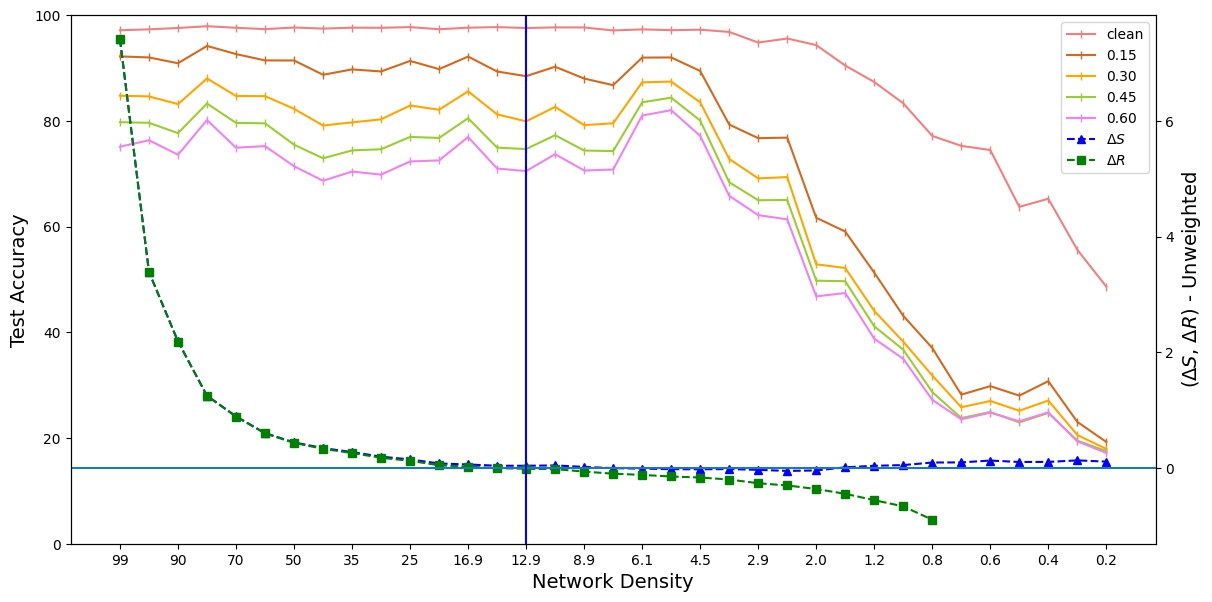}}
\centerline{(b) $W_{hh}$}
\caption{Variation in test set accuracy (left vertical axis) and spectral gap ($\Delta_S$, $\Delta_R$) (right vertical axis) on sequence MNIST data for RNN with remaining edges percentage $q$, considering unweighted graph representation. The vertical lines shows the first zero crossing of $\Delta_S$, $\Delta_R$.}
\label{fig:RNN_M:MNIST}
\end{center}
\vskip -0.2in
\end{figure}

\begin{figure}[htbp]
\begin{center}
\centerline{\includegraphics[width=\columnwidth]{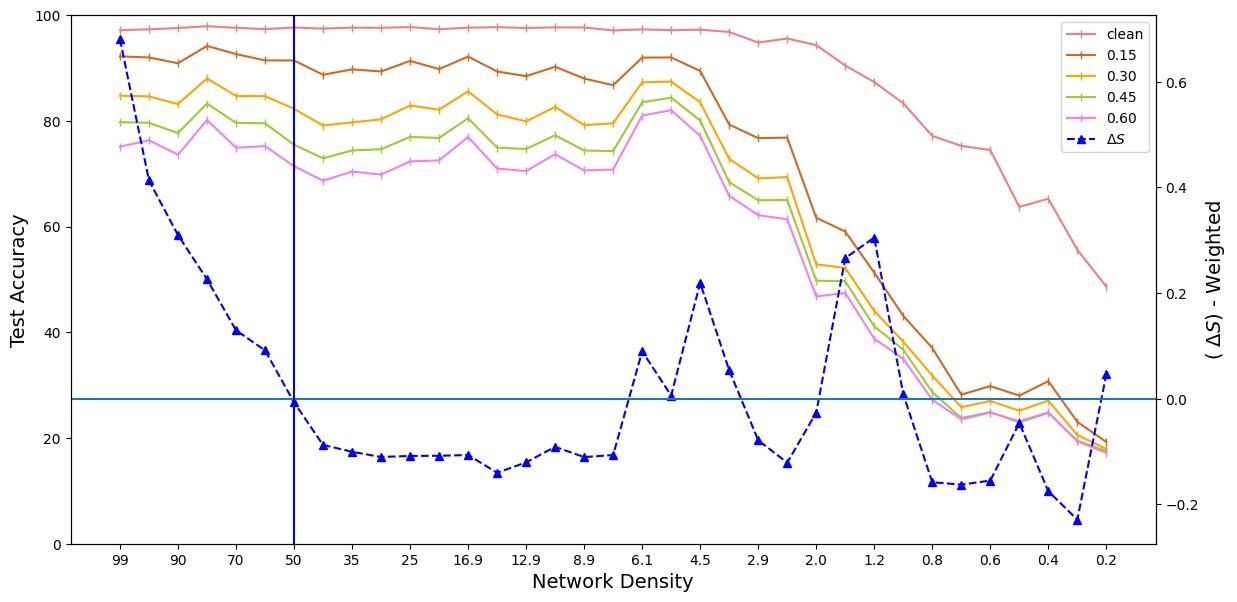}}
\centerline{(a) $W_{xh}$}
\centerline{\includegraphics[width=\columnwidth]{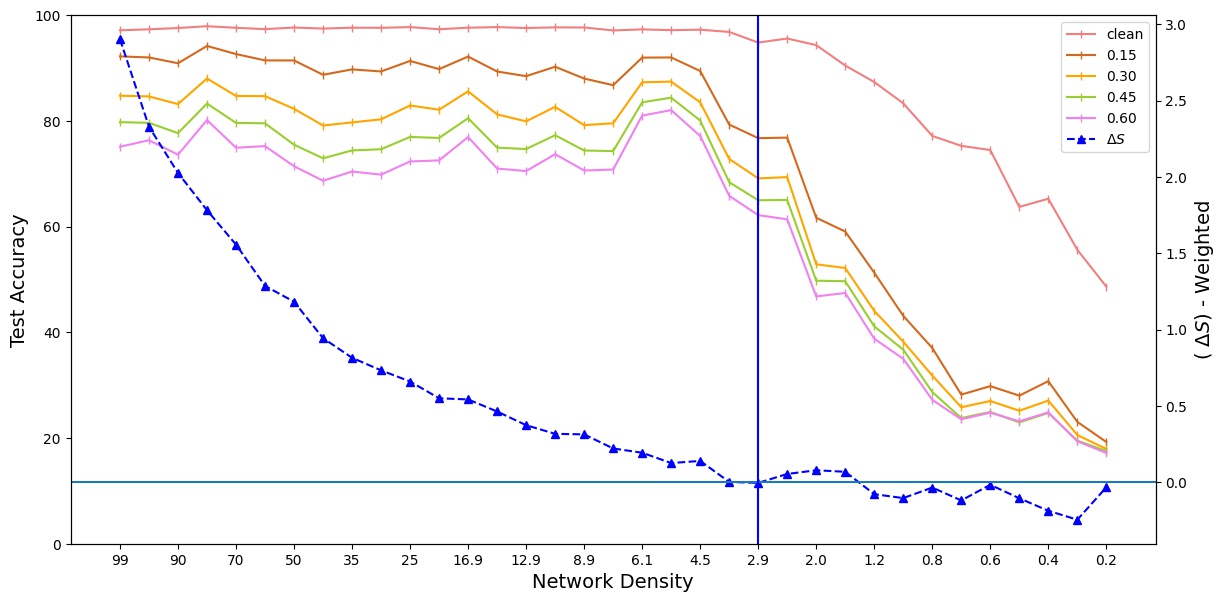}}
\centerline{(b) $W_{hh}$}
\caption{Variation in test set accuracy and spectral gap ($\Delta_S$) on sequence MNIST data for RNN with remaining edges percentage, considering weighted graph representation. The vertical line shows the first zero crossing of $\Delta_S$.}
\label{fig:RNN_W:MNIST}
\end{center}
\vskip -0.2in
\end{figure}

\begin{figure}[htbp]
\begin{center}
\centerline{\includegraphics[width=\columnwidth]{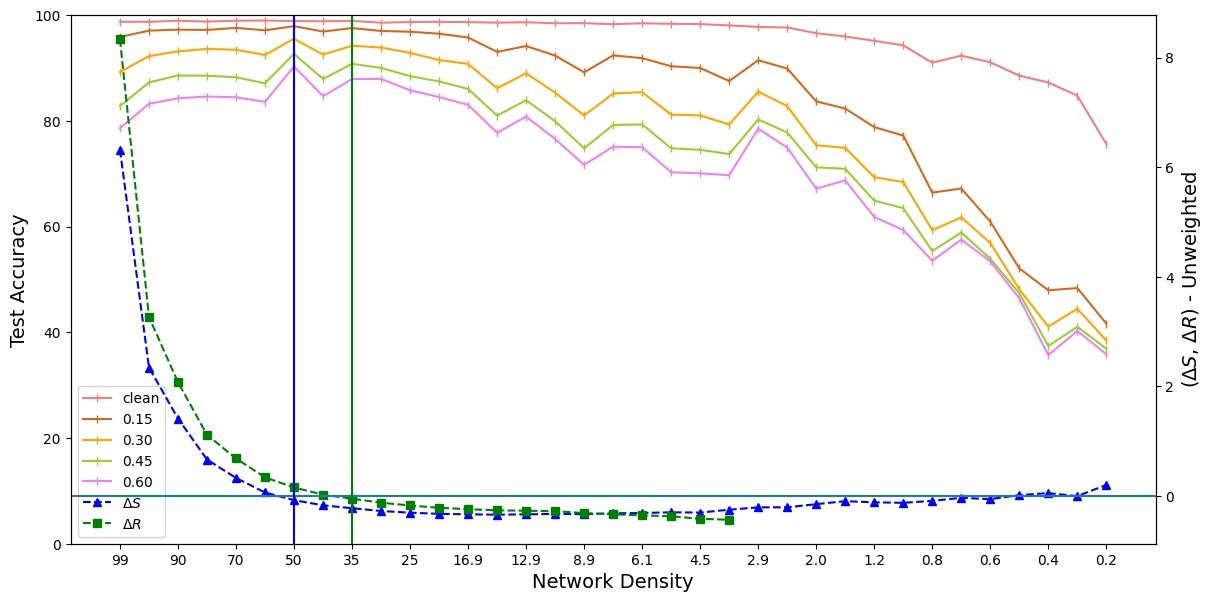}}
\centerline{(a) $W_{xh}$}
\centerline{\includegraphics[width=\columnwidth]{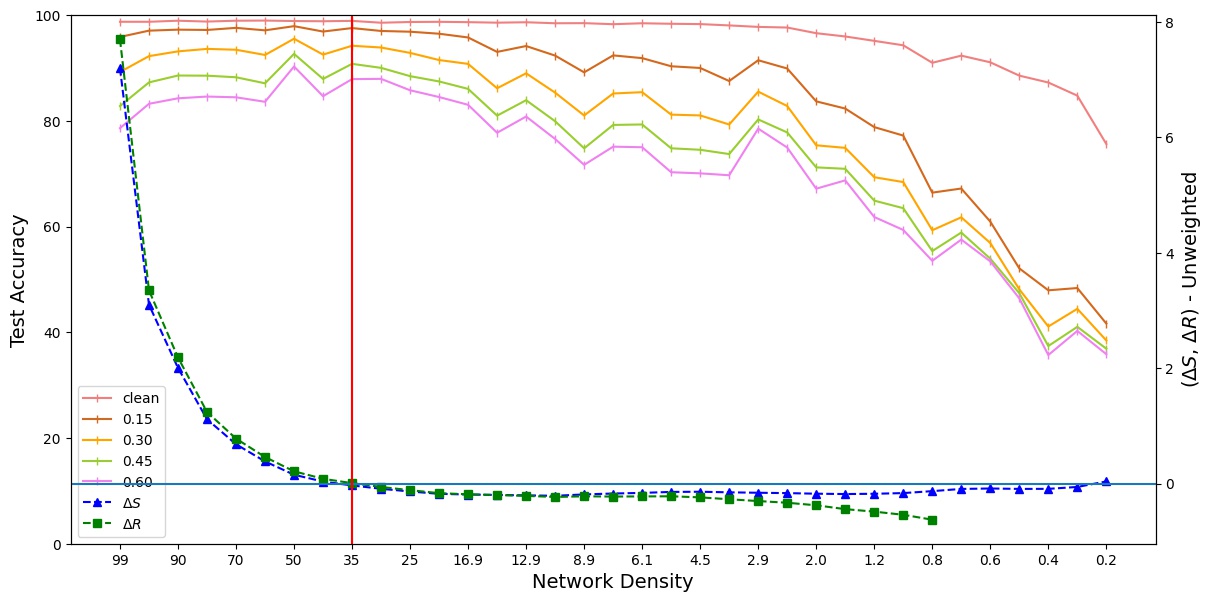}}
\centerline{(b) $W_{hh}$}
\caption{Variation in test set accuracy and spectral gap ($\Delta_S$, $\Delta_R$) on sequence MNIST data for LSTM with remaining edges percentage, considering unweighted graph representation. The vertical lines shows the first zero crossing of $\Delta_S$, $\Delta_R$.}
\label{fig:LSTM_M:MNIST}
\end{center}
\vskip -0.2in
\end{figure}

\begin{figure}[htbp]
\begin{center}
\centerline{\includegraphics[width=\columnwidth]{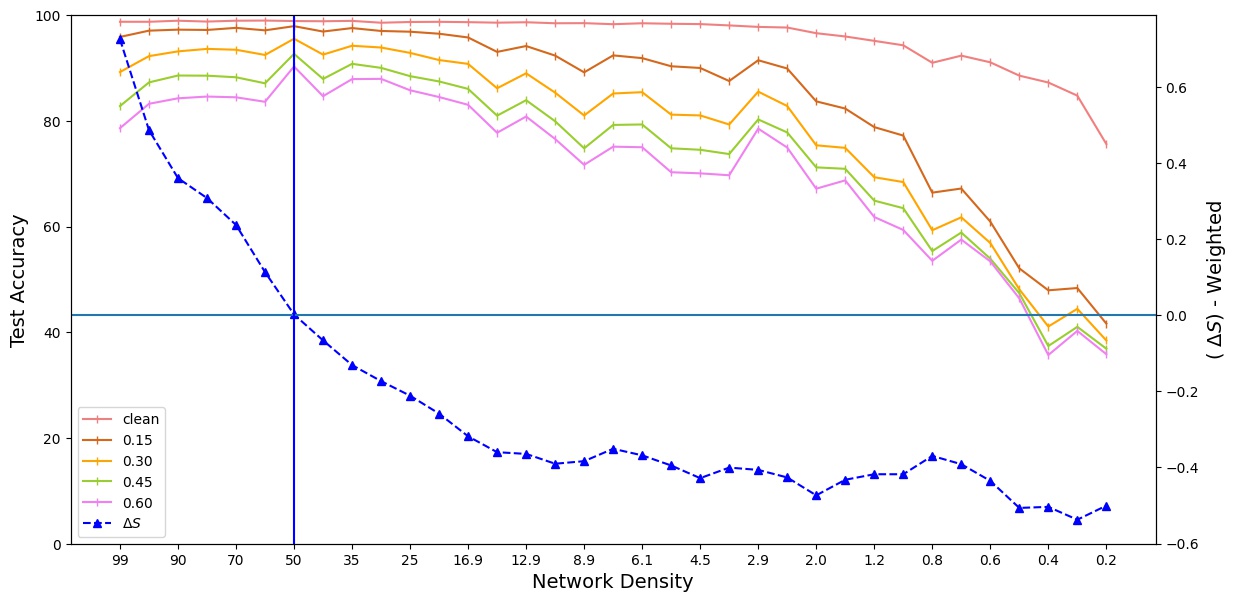}}
\centerline{(a) $W_{xh}$}
\centerline{\includegraphics[width=\columnwidth]{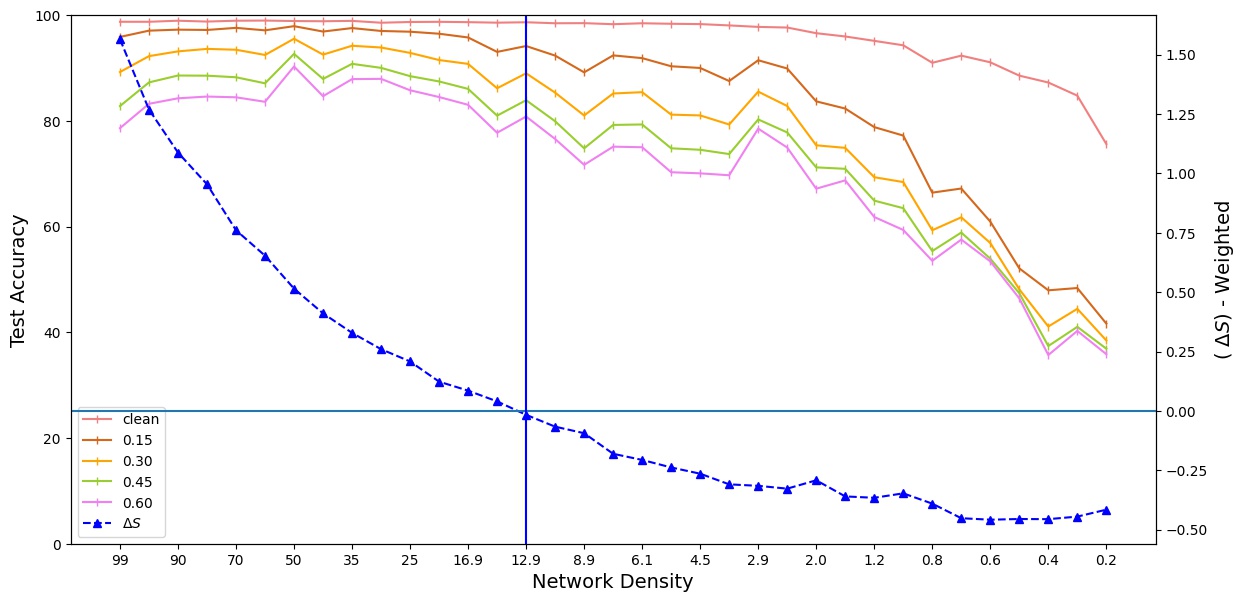}}
\centerline{(b) $W_{hh}$}
\caption{Variation in test set accuracy and spectral gap ($\Delta_S$) on sequence MNIST data for LSTM with remaining edges percentage, considering weighted representation. The vertical line shows the first zero crossing of $\Delta_S$.}
\label{fig:LSTM_W:MNIST}
\end{center}
\vskip -0.2in
\end{figure}

Figures~\ref{fig:RNN_M:MNIST} and \ref{fig:RNN_W:MNIST} shows the variation for RNN considering the unweighted and weighted adjacency matrices respectively for the MNIST dataset. Similar plots are shown for LSTM in Figures~\ref{fig:LSTM_M:MNIST} and \ref{fig:LSTM_W:MNIST}. Both clean and noisy input sequences are considered in the plots. 

\begin{figure}[htbp]
\begin{center}
\centerline{\includegraphics[width=\columnwidth]{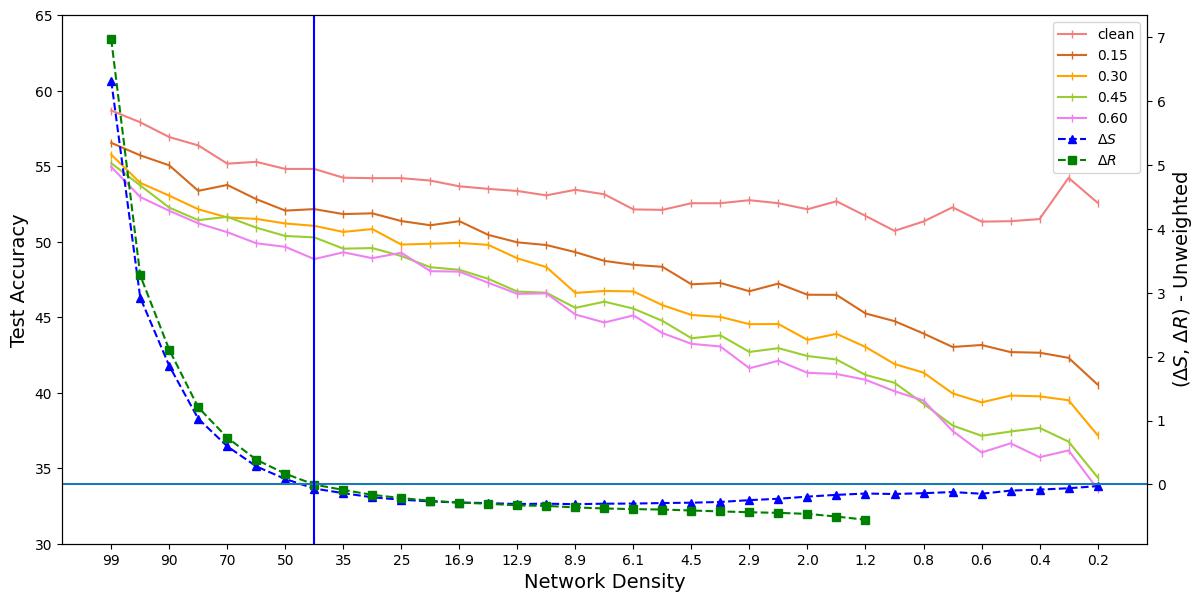}}
\centerline{(a) $W_{xh}$}
\centerline{\includegraphics[width=\columnwidth]{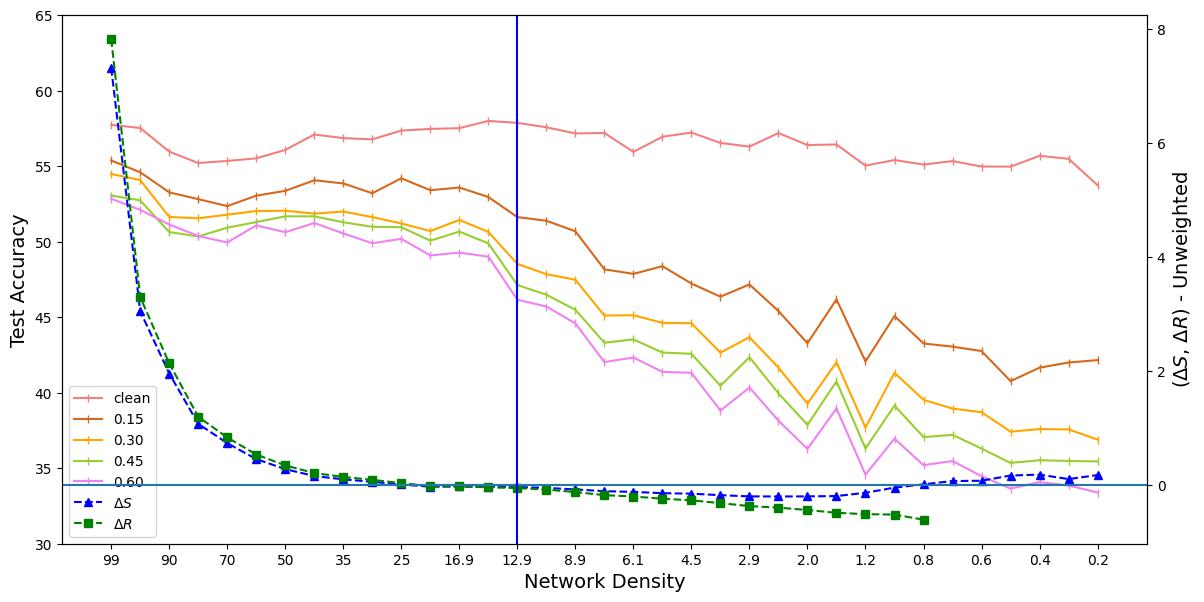}}
\centerline{(b) $W_{hh}$}
\caption{Variation in test set accuracy and spectral gap ($\Delta_S$, $\Delta_R$) on sequence CIFAR-10 data for LSTM with remaining edges percentage, considering unweighted graph representation. The vertical lines shows the first zero crossing of $\Delta_S$, $\Delta_R$.}
\label{fig:LSTM_M:CIFAR10}
\end{center}
\vskip -0.2in
\end{figure}

\begin{figure}[htbp]
\begin{center}
\centerline{\includegraphics[width=\columnwidth]{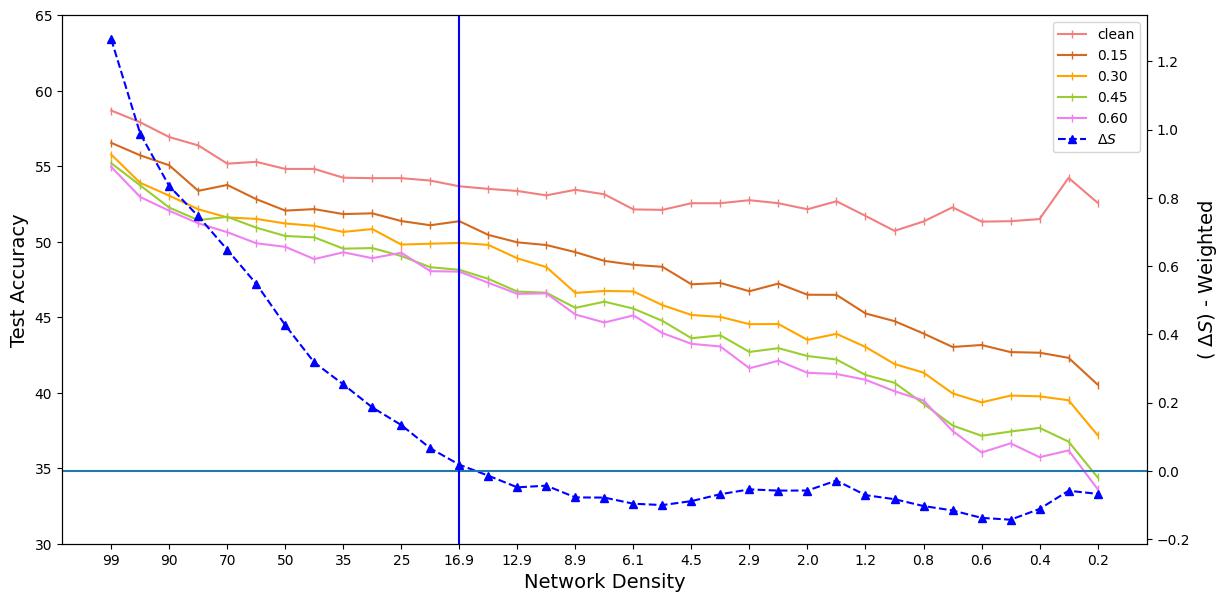}}
\centerline{(a) $W_{xh}$}
\centerline{\includegraphics[width=\columnwidth]{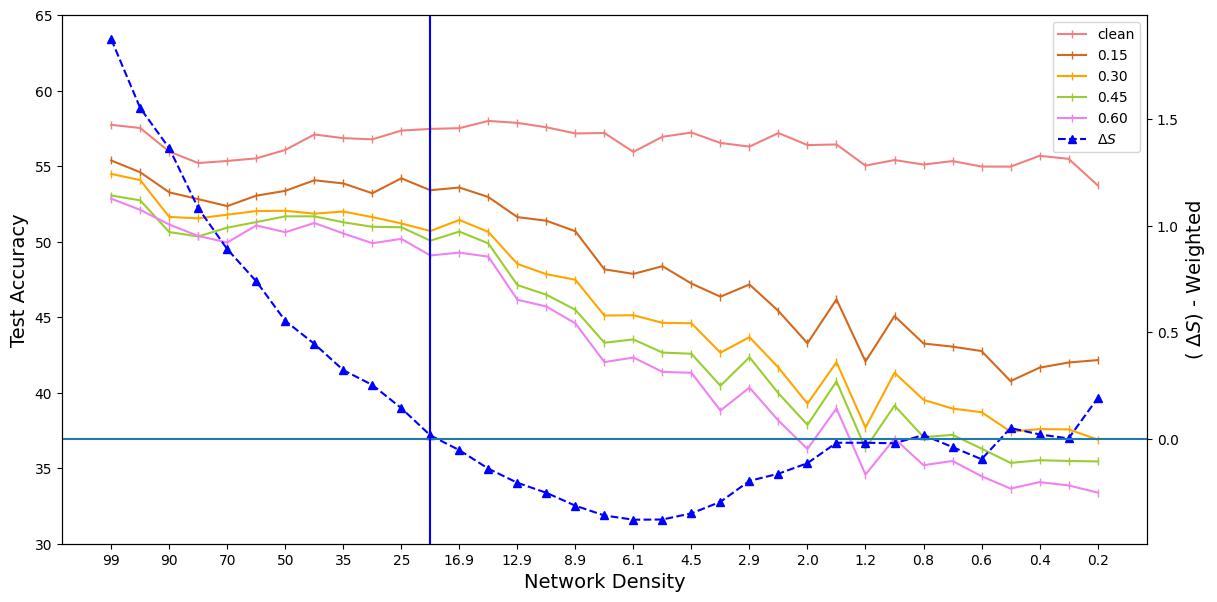}}
\centerline{(b) $W_{hh}$}
\caption{Variation in test set accuracy and spectral gap ($\Delta_S$) on sequence CIFAR-10 data for LSTM with remaining edges percentage, considering weighted representation. The vertical line shows the first zero crossing of $\Delta_S$.}
\label{fig:LSTM_W:CIFAR10}
\end{center}
\vskip -0.2in
\end{figure}

\begin{figure}[htbp]
\begin{center}
\centerline{\includegraphics[width=\columnwidth]{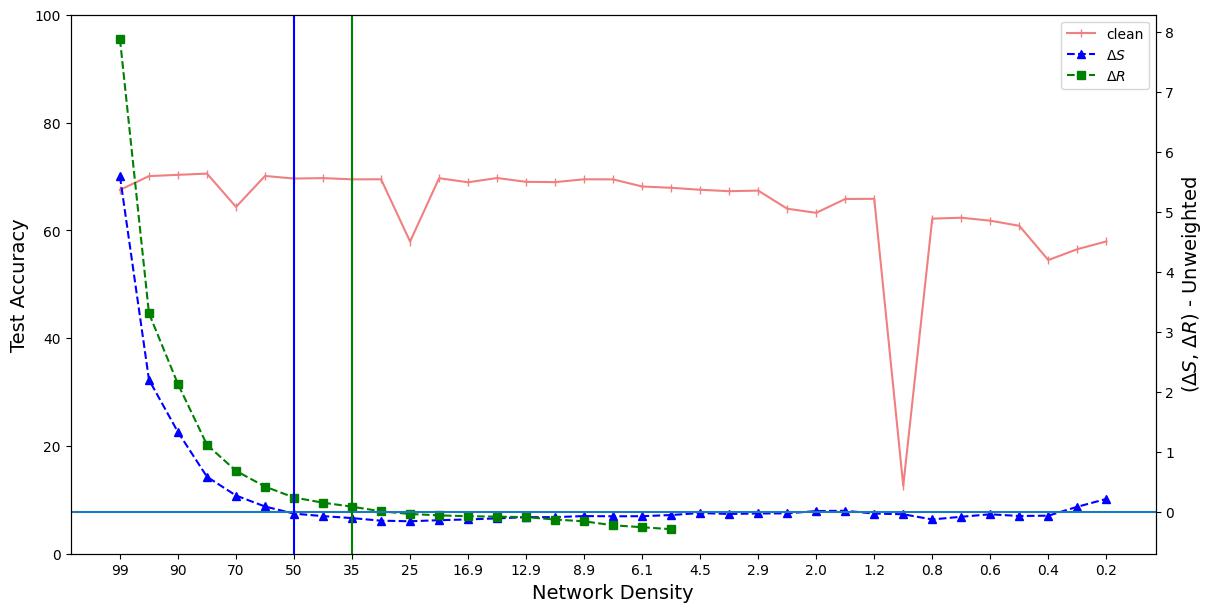}}
\centerline{(a) $W_{xh}$}
\centerline{\includegraphics[width=\columnwidth]{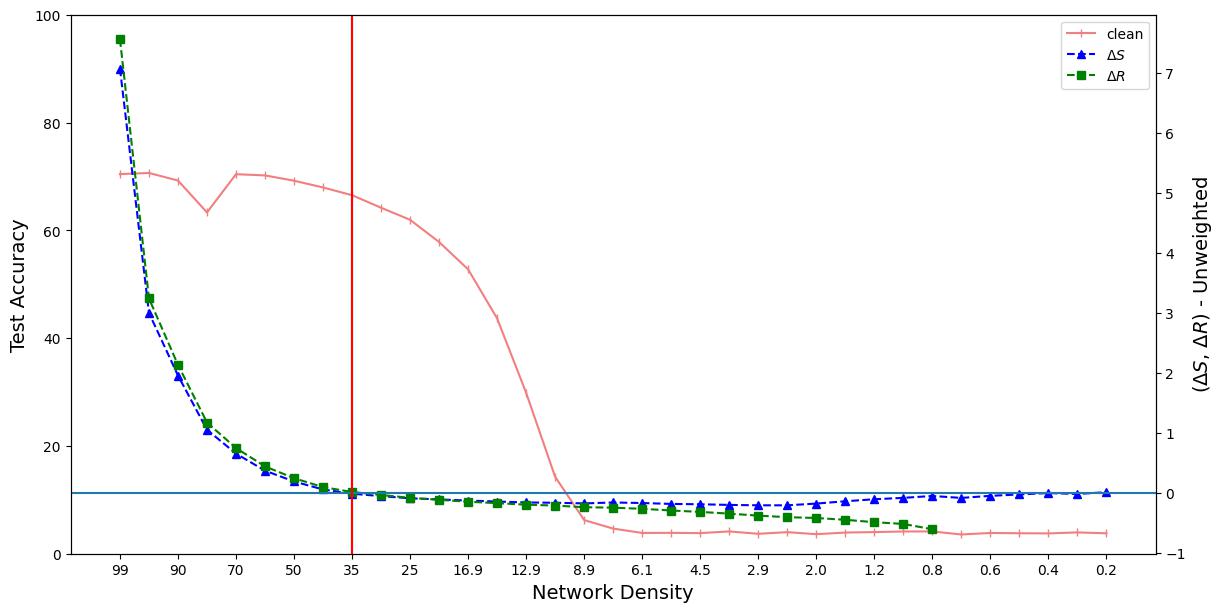}}
\centerline{(b) $W_{hh}$}
\caption{Variation in test set accuracy and spectral gap ($\Delta_S$, $\Delta_R$) on Google speech command data for LSTM with remaining edges percentage, considering unweighted graph representation. The vertical lines shows the first zero crossing of $\Delta_S$, $\Delta_R$.}
\label{fig:LSTM_M:Speech}
\end{center}
\vskip -0.2in
\end{figure}

\begin{figure}[htbp]
\begin{center}
\centerline{\includegraphics[width=\columnwidth]{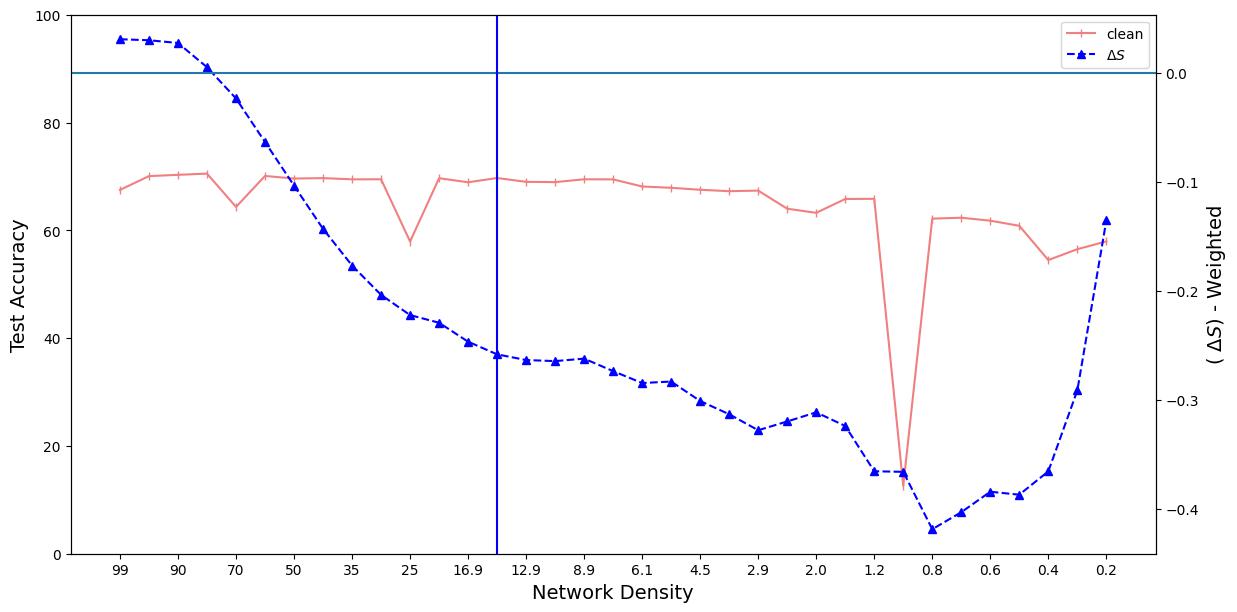}}
\centerline{(a) $W_{xh}$}
\centerline{\includegraphics[width=\columnwidth]{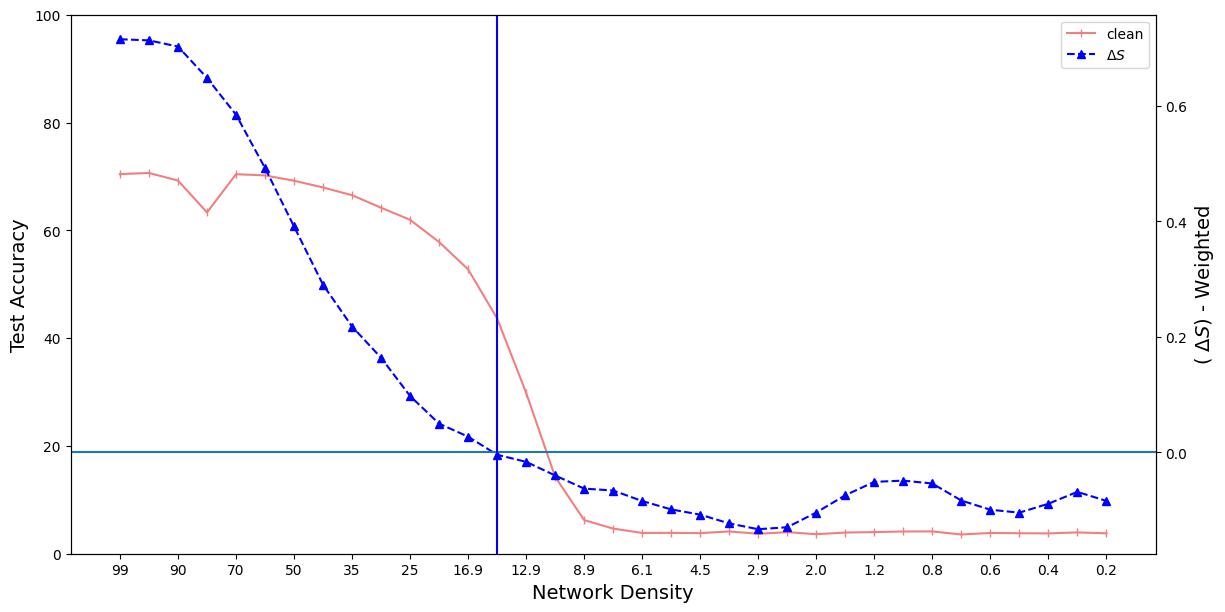}}
\centerline{(b) $W_{hh}$}
\caption{Variation in test set accuracy and spectral gap ($\Delta_S$) on Google speech command data for LSTM with remaining edges percentage, considering weighted representation. The vertical line shows the first zero crossing of $\Delta_S$.}
\label{fig:LSTM_W:Speech}
\end{center}
\vskip -0.2in
\end{figure}

The variation of spectral gaps for the CIFAR10 data for LSTM is shown in Figures~\ref{fig:LSTM_M:CIFAR10} and \ref{fig:LSTM_W:CIFAR10} for unweighted and weighted graph representations. Next, we consider the speech command dataset with much longer sequence length of $k=400$. The variation in spectral gaps for LSTM are shown in Figures~\ref{fig:LSTM_M:Speech} and \ref{fig:LSTM_W:Speech} for unweighted and weighted graphs.

It is observed from the figures that as long as the expander graph property holds for the networks the test set classification accuracy is almost preserved as compared to the unpruned network. Whereas the accuracy starts dropping when these properties are lost. 

For the MNIST dataset, in the RNN, the $W_{xh}$ weights lose the Ramanujan property at a remaining edge percentage of $50.0$\%, and the $W_{hh}$ weights lose the property at $50.0$\% and $35.0$\% for the unweighted graph representation (Figure~\ref{fig:RNN_M:MNIST}) and at $2.9$\% remaining weights (Figure~\ref{fig:RNN_W:MNIST}) considering the weighted graph representation. For LSTM with the MNIST dataset, with an unweighted representation, the remaining weights are $50.0$\% ($W_{xh}$) and $35.0$\% ($W_{hh}$) (Figure~\ref{fig:LSTM_M:MNIST}) when the expander graph property is lost. The $W_{xh}$ weights lose the Ramanujan property at a remaining edge percentage of $50.0$\%, and the $W_{hh}$ weights lose the property at $12.9$\% remaining weights (Figure~\ref{fig:LSTM_W:MNIST}) for the weighted graph representation, for the same dataset,

LSTM on the CIFAR-10 datset has a zero crossing of spectral gaps at a remaining edge percentage of $47.0$\% ($W_{xh}$) and $12.9$\% ($W_{hh}$) for the unweighted representation (Figure~\ref{fig:LSTM_M:CIFAR10}). The values are $16.9$\%, for $W_{xh}$ and $18.0$\% for $W_{hh}$ (Figure~\ref{fig:LSTM_W:CIFAR10}) for the weighted graph representation.  For the Google speech command dataset, having a longer sequence length, we observe that the zero crossing for the unweighted representation is at remaining weight percentage of $50$\% ($W_{xh}$) and $35$\% ($W_{hh}$) for the unweighted representation (Figure~\ref{fig:LSTM_M:Speech}), and $14.9$\% for both $W_{xh}$ and $W_{hh}$ in the weighted graph representation (Figure~\ref{fig:LSTM_W:Speech}).

We have added various noise quantities to the MNIST and CIFAR-10 dataset as described in Section~\ref{sec:data}. The Google speech command data contains natural background noise. We observe from Figures~\ref{fig:RNN_M:MNIST}-\ref{fig:LSTM_W:CIFAR10} that the degradation in performance with loss of expander graph property becomes even more prominent as noise quantity increases. This reinstates the fact that expander graph property is crucial for noise robustness of the networks.

We observe in most of our experiments with RNN that the $W_{xh}$ layer lose the expander graph property before the $W_{hh}$ layer. This points to the fact that $W_{xh}$ edges play more significant role as compared to the $W_{hh}$. Similar, results has been reported in \cite{averbeck23}.

The value of $\Delta_S$ oscillates for RNN (Figure~\ref{fig:RNN_W:MNIST}) for very sparse networks. This may be due the approximation errors for very sparse networks. Other oscillations of test accuracy at lower lesser sparsity may be attributed to underfitting or overfitting that we do not optimize.

\section{Conclusion}
In this paper we study the expansion properties of sparse RNN and LSTM networks obtained by pruning. We empirically observe that the sparse networks that satisfy the expander graph properties preserve the classification accuracy of the unpruned network. We also show using results from linear algebra that bounds on the spectral gaps for recurrent network structures can be obtained from the corresponding bounds for the layer bi-partite graphs.  We present results for the sequential MNIST, CIFAR-10, and the Google speech command data with and without noise for both RNN and LSTM pruned networks using iterative magnitude pruning to support our claims. 

\bibliography{iclr2024_conference}
\bibliographystyle{iclr2024_conference}

\end{document}

%% file: math_commands.tex
\usepackage{amsmath,amsfonts,bm}

\def\eqref#1{equation~\ref{#1}}

\def\1{\bm{1}}

\def\ra{{\textnormal{a}}}

\def\rx{{\textnormal{x}}}

\def\rva{{\mathbf{a}}}

\def\erva{{\textnormal{a}}}

\def\ervx{{\textnormal{x}}}

\def\rmA{{\mathbf{A}}}

\def\vmu{{\bm{\mu}}}
\def\vtheta{{\bm{\theta}}}
\def\va{{\bm{a}}}

\def\ve{{\bm{e}}}

\def\vx{{\bm{x}}}

\def\eva{{a}}

\def\mA{{\bm{A}}}

\def\mH{{\bm{H}}}
\def\mI{{\bm{I}}}
\def\mJ{{\bm{J}}}

\def\mX{{\bm{X}}}

\def\mSigma{{\bm{\Sigma}}}

\DeclareMathAlphabet{\mathsfit}{\encodingdefault}{\sfdefault}{m}{sl}
\SetMathAlphabet{\mathsfit}{bold}{\encodingdefault}{\sfdefault}{bx}{n}
\newcommand{\tens}[1]{\bm{\mathsfit{#1}}}
\def\tA{{\tens{A}}}

\def\tX{{\tens{X}}}

\def\gG{{\mathcal{G}}}

\def\sA{{\mathbb{A}}}
\def\sB{{\mathbb{B}}}

\def\sS{{\mathbb{S}}}

\def\emA{{A}}

\newcommand{\etens}[1]{\mathsfit{#1}}

\def\etA{{\etens{A}}}

\newcommand{\E}{\mathbb{E}}

\newcommand{\R}{\mathbb{R}}

\newcommand{\KL}{D_{\mathrm{KL}}}
\newcommand{\Var}{\mathrm{Var}}

\newcommand{\Cov}{\mathrm{Cov}}

\newcommand{\normltwo}{L^2}
\newcommand{\normlp}{L^p}

\newcommand{\parents}{Pa} 